\pdfoutput=1

\documentclass{article}

\usepackage[final]{nips_2016}

\usepackage[utf8]{inputenc} 
\usepackage[T1]{fontenc}    
\usepackage{hyperref}       
\usepackage{url}            
\usepackage{booktabs}       
\usepackage{amsfonts}       
\usepackage{nicefrac}       
\usepackage{microtype}      
\usepackage{amsmath}
\usepackage{graphicx}
\usepackage{url}

\title{Combining Fully Convolutional and Recurrent Neural Networks for 3D Biomedical Image Segmentation}

\author{
  Jianxu Chen \\
  University of Notre Dame\\
  \texttt{jchen16@nd.edu} \\
  \And
  Lin Yang \\
 University of Notre Dame \\
 \texttt{lyang5@nd.edu} \\
  \AND
  Yizhe Zhang \\
  University of Notre Dame \\
  \texttt{yzhang29@nd.edu} \\
 \And
  Mark Alber \\
  University of Notre Dame \\
  \texttt{malber@nd.edu} \\
 \And
 Danny Z. Chen \\
  University of Notre Dame\\
  \texttt{dchen@nd.edu} \\
}

\begin{document}

\maketitle

\begin{abstract}

Segmentation of 3D images is a fundamental problem in biomedical image analysis. Deep learning (DL) approaches have achieved state-of-the-art segmentation performance. To exploit the 3D contexts using neural networks, known DL segmentation methods, including 3D convolution, 2D convolution on planes orthogonal to 2D image slices, and LSTM in multiple directions, all suffer incompatibility with the highly anisotropic dimensions in common 3D biomedical images. In this paper, we propose a new DL framework for 3D image segmentation, based on a combination of a fully convolutional network (FCN) and a recurrent neural network (RNN), which are responsible for exploiting the intra-slice and inter-slice contexts, respectively. To our best knowledge, this is the first DL framework for 3D image segmentation that explicitly leverages 3D image anisotropism. Evaluating using a dataset from the ISBI Neuronal Structure Segmentation Challenge and in-house image stacks for 3D fungus segmentation, our approach achieves promising results comparing to the known DL-based 3D segmentation approaches.    

\end{abstract}

\section{Introduction}
\label{sec:intro}

In biomedical image analysis, a fundamental problem is the segmentation of 3D images, to identify target 3D objects such as neuronal structures \cite{ISBI_EM} and knee cartilage \cite{triplane}. In biomedical imaging, 3D images often consist of highly anisotropic dimensions \cite{2d3d}, that is, the scale of each voxel in depth (the $z$-axis) can be much larger (e.g., $5\!\sim\!10$ times) than that in the $xy$ plane.

On various biomedical image segmentation tasks, deep learning (DL) methods have achieved tremendous success in terms of accuracy (outperforming classic methods by a large margin \cite{IDSIA}) and generality (mostly application-independent \cite{unet}). For 3D segmentation, known DL schemes can be broadly classified into four categories. (I) 2D fully convolutional networks (FCN), such as U-Net \cite{unet} and DCAN \cite{dcan}, can be applied to each 2D image slice, and 3D segmentation is then generated by concatenating the 2D results. (II) 3D convolutions can be employed to replace 2D convolutions \cite{3Dconv}, or combined with 2D convolutions into a hybrid network \cite{2d3d}. (III) Tri-planar schemes (e.g., \cite{triplane}) apply three 2D convolutional networks based on orthogonal planes (i.e., the $xy$, $yz$, and $xz$ planes) to perform voxel classification. (IV) 3D segmentation can also be conducted by recurrent neural networks (RNN). A most representative RNN based scheme is Pyramid-LSTM \cite{pyramid}, which uses six generalized long short term memory networks to exploit the 3D context.

There are mainly three issues to the known DL-based 3D segmentation methods. First, simply linking 2D segmentations into 3D cannot leverage the spatial correlation along the $z$-direction. Second, incorporating 3D convolutions may incur extremely high computation costs (e.g., high memory consumption and long training time \cite{3Dconv}). Third, both 3D convolution and other circumventive solutions (to reduce intensive computation of 3D convolution), like tri-planar schemes or Pyramid-LSTM, perform 2D convolutions with isotropic kernel on anisotropic 3D images. This could be problematic, especially for images with substantially lower resolution in depth (the $z$-axis). For instance, both the tri-planar schemes and Pyramid-LSTM perform 2D convolutions on the $xz$ and $yz$ planes. For two orthogonal one-voxel wide lines in the $xz$ plane, one along the $z$-direction and the other along the $x$-direction, they may correspond to two structures at very different scales, and consequently may correspond to different types of objects --- or even may not both correspond to objects of interest. But, 2D convolutions on the $xz$ plane with isotropic kernel are not able to differentiate these two lines. On the other hand, 3D objects of a same type, if rotated in 3D, may have very different appearances in the $xz$ or $yz$ plane. This fact makes the features extracted by such 2D isotropic convolutions in the $xz$ or $yz$ plane suffer poor generality (e.g., may cause overfitting). 

In common practice, a 3D biomedical image is often represented as a sequence of 2D slices (called a $z$-stack). Recurrent neural networks, especially LSTM \cite{LSTM}, are an effective model to process sequential data \cite{convLSTM,Shi}. Inspired by these facts, we propose a new framework combining two DL components: a fully convolutional network (FCN) to extract intra-slice contexts, and a recurrent neural network (RNN) to extract inter-slice contexts. Our framework is based on the following ideas.

Our FCN component employs a new deep architecture for 2D feature extraction. It aims to efficiently compress the intra-slice information into hierarchical features. Comparing to known FCN for 2D biomedical imaging (e.g., U-Net \cite{unet}), our new FCN is considerably more effective in dealing with objects of very different scales by simulating human behaviors in perceiving multi-scale information.

We introduce a generalized RNN to exploit 3D contexts, which essentially applies a series of 2D convolutions on the $xy$ plane in a recurrent fashion to interpret 3D contexts while propagating contextual information in the $z$-direction. Our key idea is to hierarchically assemble intra-slice contexts into 3D contexts by leveraging the inter-slice correlations. The insight is that our RNN can distill 3D contexts in the same spirit as the 2D convolutional neural network (CNN) extracting a hierarchy of contexts from a 2D image. Comparing to known RNN models for 3D segmentation, such as Pyramid-LSTM \cite{pyramid}, our RNN model is free of the problematic isotropic convolutions on anisotropic images, and can exploit 3D contexts more efficiently by combining with FCN.    


The essential difference between our new DL framework and the known DL-based 3D segmentation approaches is that we explicitly leverage the anisotropism of 3D images and efficiently construct a hierarchy of discriminative features from 3D contexts by performing systematic 2D operations. Our framework can serve as a new paradigm of migrating 2D DL architectures (e.g., CNN) to effectively exploit 3D contexts and solve 3D image segmentation problems.

\begin{figure}[tb] 
   \centering
   \includegraphics[width=4.0in]{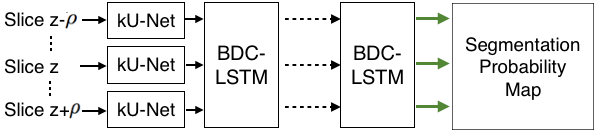} 
   \caption{An overview of our DL framework for 3D segmentation. There are two key components in the architecture: $k$U-Net and BDC-LSTM. $k$U-Net is a type of FCN and is applied to 2D slices to exploit intra-slice contexts. BDC-LSTM, a generalized LSTM network, is applied to a sequence of 2D feature maps, from 2D slice $z-\rho$ to 2D slice $z+\rho$, extracted by $k$U-Nets, to extract hierarchical features from the 3D contexts. Finally, a softmax function (the green arrows) is applied to the result of each slice in order to build the segmentation probability map.}
   \label{fig:overview}
\end{figure}

\section{Methodology}

A schematic view of our DL framework is given in Fig.~\ref{fig:overview}. This framework is a combination of two key components: an FCN (called $k$U-Net) and an RNN (called BDC-LSTM), to exploit intra-slice and inter-slice contexts, respectively. Section~\ref{sec:kunet} presents the $k$U-Net, and Section~\ref{sec:rnn} introduces the derivation of the BDC-LSTM. We then show how to combine these two components in the framework to conduct 3D segmentation. Finally, we discuss the training strategy.  

\begin{figure}[tb]
   \centering
   \begingroup
      \leftskip -2em
     \includegraphics[width = 6in]{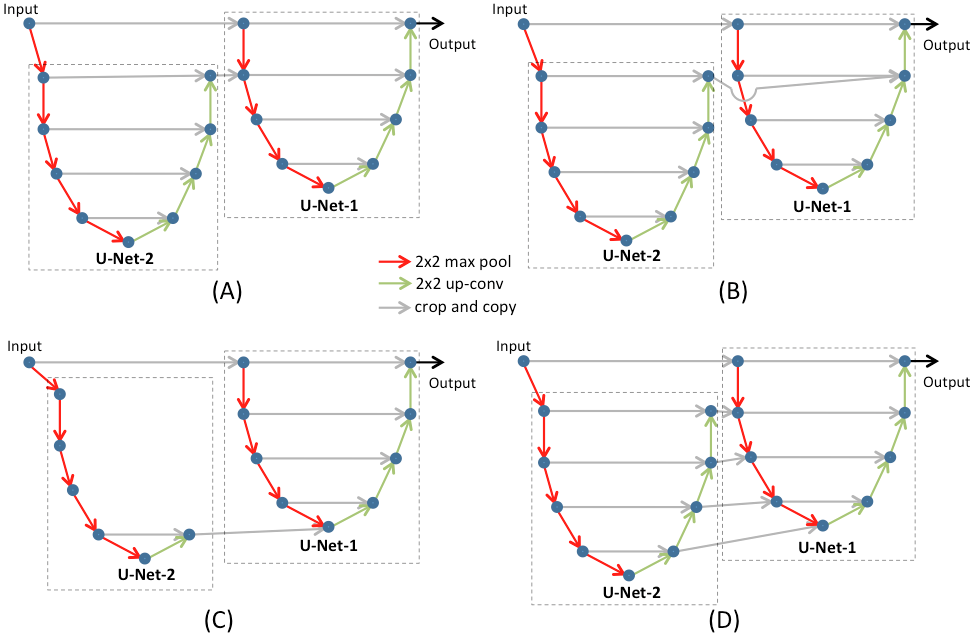} 
   \caption{Illustrating four different ways to organize $k$ submodule U-Nets in $k$U-Net (here $k=2$). U-Net-$2$ works in a coarser scale (downsampled once from the original image), while U-Net-$1$ works in a finer scale (directly cropped from the original image). $k$U-Net propagates high level information extracted by U-Net-2 to U-Net-1. (A) U-Net-$1$ fuses the output of U-Net-$2$ in the downsampling stream. (B) U-Net-$1$ fuses the output of U-Net-$2$ in the upsampling stream. (C) U-Net-$1$ fuses the intermediate result of U-Net-$2$ in the most abstract layer. (D) U-Net-$1$ takes every piece of information from U-Net-$2$ in the commensurate layers. Architecture (A) is finally adopted for $k$U-Net.}
   \label{fig:kunet}
   \endgroup
\end{figure}

\subsection{The FCN Component: $k$U-Net}
\label{sec:kunet}

The FCN component aims to construct a feature map for each 2D slice, from which object-relevant information (e.g., texture, shapes) will be extracted and object-irrelevant information (e.g., uneven illumination, imaging contrast) will be discarded. By doing so, the next RNN component can concentrate on the inter-slice context. 

A key challenge to the FCN component is the multi-scale issue. Namely, objects in biomedical images, specifically in 2D slices, can have very different scales and shapes. But, the common FCN \cite{FCN} and other known variants for segmenting biomedical images (e.g., U-Net \cite{unet}) work on a fixed-size perception field (e.g., a $500\times500$ region in the whole 2D slice). When objects are of larger scale than the pre-defined perception field size, it can be troublesome for such FCN methods to capture the high level context (e.g., the overall shapes). In the literature, a multi-stream FCN was proposed in ProNet \cite{pronet} to address this multi-scale issue in natural scene images. In ProNet, the same image is resized to different scales and fed in parallel to a shared FCN with the same parameters. However, the mechanism of shared parameters may make it not suitable for biomedical images, because objects of different scales may have very different appearances and require different FCNs to process. 

We propose a new FCN architecture to simulate how human experts perceive multi-scale information, in which multiple submodule FCNs are employed to work on different image scales systematically. Here, we use U-Net \cite{unet} as the submodule FCN and call the new architecture $k$U-Net. U-Net \cite{unet} is chosen because it is a well-known FCN achieving huge success in biomedical image segmentation. U-Net \cite{unet} consists of four downsampling steps followed by four upsampling steps. Skip-layer connections exist between each downsampled feature map and the commensurate upsampled feature map. We refer to \cite{unet} for the detailed structure of U-Net.

We observed that, when human experts label the ground truth, they tend to first zoom out the image to figure out where are the target objects and then zoom in to label the accurate boundaries of those targets. There are two critical mechanisms in $k$U-Net to simulate such human behaviors. (1) $k$U-Net employs a sequence of submodule FCNs to extract information at different scales sequentially (from the coarsest scale to the finest scale). (2) The information extracted by the submodule FCN responsible for a coarser scale will be propagated to the subsequent submodule FCN to assist the feature extraction in a finer scale. 

First, we create different scales of an original input 2D image by a series of connections of $k-1$ max-pooling layers. Let $I_t$ be the image of scale $t$ ($t=1,\ldots, k$), i.e., the result after $t-1$ max-pooling layers ($I_1$ is the original image). Each pixel in $I_t$ corresponds to $2^{t-1}$ pixels in the original image. Then, we use U-Net-$t$ ($t=1,\ldots, k$), i.e., the $t$-th submodule, to process $I_{t}$. We keep the input window size the same across all U-Nets by using crop layers. Intuitively, U-Net-$1$ to U-Net-$k$ all have the same input size, while U-Net-$1$ views the smallest region with the highest resolution and U-Net-$k$ views the largest region with the lowest resolution. In other words, for any $1\leq t_1<t_2\leq k$, U-Net-$t_2$ is responsible for a larger image scale than U-Net-$t_1$.

Second, we need to propagate the higher level information extracted by U-Net-$t$ ($2\leq t\leq k$) to the next submodule, i.e., U-Net-($t-1$), so that clues from a coarser scale can assist the work in a finer scale. A natural strategy is to copy the result from U-Net-$t$ to the commensurate layer in U-Net-($t-1$). As shown in Fig.~\ref{fig:kunet}, there are four typical ways to achieve this: (A) U-Net-($t-1$) only uses the final result from U-Net-$t$ and uses it at the start; (B) U-Net-($t-1$) only uses the final result from U-Net-$t$ and uses it at the end; (C) U-Net-($t-1$) only uses the most abstract information from U-Net-$t$; (D) U-Net-($t-1$) uses every piece of information from U-Net-$t$. Based on our trial studies, type (A) and type (D) achieved the best performance. Since type (A) has fewer parameters than (D), we chose type (A) as our final architecture to organize the sequence of submodule FCNs.

From a different perspective, each submodule U-Net can be viewed as a ``super layer". Therefore, the $k$U-Net is a ``deep'' deep learning model. Because the parameter $k$ exponentially increases the input window size of the network, a small $k$ is sufficient to handle many biomedical images (we use $k=2$ in our experiments). Appended with a $1\!\times\!1$ convolution (to convert the number of channels in the feature map) and a softmax layer, the $k$U-Net can be used for 2D segmentation problems. We will show (see Table~\ref{tab:exp}) that $k$U-Net (i.e., a sequence of collaborative U-Nets) can achieve better performance than a single U-Net in terms of segmentation accuracy.

\subsection{The RNN Component: BDC-LSTM}
\label{sec:rnn}

In this section, we first review the classic LSTM network \cite{LSTM}, and sketch how to generalize the classic LSTM to convolutional LSTM \cite{convLSTM,Shi,pyramid} (denoted by CLSTM). Next, we describe how our RNN component, called BDC-LSTM, is extended from CLSTM. Finally, we propose a deep architecture for BDC-LSTM, and discuss its advantages over other variants.

\textbf{Classic LSTM}: RNN (e.g., LSTM) is a neural network that maintains a self-connected internal status acting as a ``memory". The ability to ``remember'' what has been seen allows RNN to attain exceptional performance in processing sequential data. Classic LSTM \cite{LSTM} can be defined as follows.

\begin{equation}
\label{equ:lstm}
\left\{
   \begin{array}{ll}
       i_t  = \sigma(x_t W_{xi} + h_{t-1}W_{hi} + b_i)\\
       f_t = \sigma(x_t W_{xf} + h_{t-1}W_{hf} + b_f)\\
       c_t = c_{t-1} f_t + i_t \;tanh(x_t  W_{xc} + h_{t-1} W_{hc} + b_c) \\
       o_t = \sigma(x_t W_{xo} + h_{t-1} W_{ho} + b_o)\\
       h_t = o_t \; tanh(c_t) 
      \end{array}
\right.
\end{equation}

Here, $\sigma()$ and $tanh()$ are logistic sigmoid and hyperbolic tangent functions; $i_t$, $f_t$, $o_t$ are the input gate, forget gate, and output gate, $b_i$, $b_f$, $b_c$, $b_o$ are bias terms, and $x_t$, $c_t$, $h_t$ are the input, the cell activation state, and the hidden state, at time $t$. $W_{**}$ are diagonal weight matrices governing the value transitions. For instance, $W_{hf}$ controls how the forget gate takes values from the hidden state.


\textbf{CLSTM}: Recently, a generalized LSTM, denoted by CLSTM, was developed \cite{convLSTM,Shi,pyramid}. CLSTM explicitly assumes that the input is images and replaces the vector multiplication in LSTM gates by convolutional operators. It is particularly efficient in exploiting image sequences. For instance, it can be used for image sequence prediction either in an encoder-decoder framework \cite{Shi} or by combining with optical flows \cite{convLSTM}. Specifically, CLSTM can be formulated as follows.


\begin{equation}
\label{equ:clstm}
\left\{
   \begin{array}{ll}
       i_z  = \sigma(x_z * W_{xi} + h_{z-1}*W_{hi} + b_i)\\
       f_z = \sigma(x_z*W_{xf} + h_{z-1}*W_{hf} + b_f)\\
       c_z = c_{z-1}\odot f_z + i_z\odot tanh(x_z * W_{xc} + h_{z-1}*W_{hc}+b_c) \\
       o_z = \sigma(x_z*W_{xo} + h_{z-1}*W_{ho} + b_o)\\
       h_z = o_z \odot tanh(c_z) 
      \end{array}
\right.
\end{equation}

Here, $*$ denotes convolution and $\odot$ denotes element-wise product. Except different operators, CLSTM has the same formulation as classic LSTM in Equ.~(\ref{equ:lstm}). We replace the index $t$ by $z$ in order to emphasize that CLSTM works slice by slice in the depth ($z$) direction. The input to CLSTM is a feature map of size $f_{in}\times l_{in} \times w_{in}$, and the output is a feature map of size $f_{out}\times l_{out} \times w_{out}$, $l_{out}\leq l_{in}$ and $w_{out}\leq w_{in}$, and $l_{out}$ and $w_{out}$ depend on the size of the convolution kernels and whether padding is used.

\begin{figure}[tb]
   \centering
   \includegraphics[width = 4.35in]{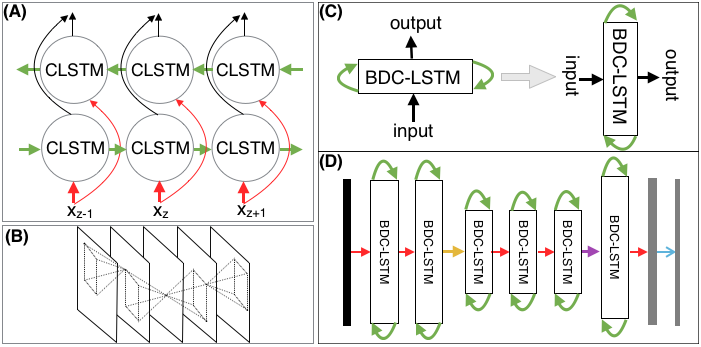} 
   \caption{(A) The structure of BDC-LSTM, where two layers of CLSTM modules are connected in a bi-directional manner. (B) A graphical illustration of information propagation through BDC-LSTM along the $z$-direction. (C) The circuit diagram of BDC-LSTM. The green arrows represent the recurrent connections in opposite directions. When rotating this diagram by 90 degrees, it has a similar structure of a layer in CNN, except the recurrent connections. (D) The deep structure of BDC-LSTM used in our method. BDC-LSTM can be stacked in a way analogous to a layer in CNN. The red arrows are $5\times5$ convolutions. The yellow and purple arrows indicate max-pooling and deconvolution, respectively. The rightmost blue arrow indicates a $1\times1$ convolution. Dropout is applied (not shown) after the input layer, the max-pooling layer and the deconvolution layer. }
   \label{fig:bdclstm}
\end{figure}

\textbf{BDC-LSTM}: We extend CLSTM to Bi-Directional Convolutional LSTM (BDC-LSTM). The key extension is to stack two layers of CLSTM, which work in two opposite directions (see Fig.~\ref{fig:bdclstm}(A)). The contextual information carried in the two layers, one in $z^{-}$-direction and the other in $z^{+}$-direction, is concatenated as output. It can be interpreted as follows. To determine the hidden state at a slice $z$, we take the 2D hierarchical features in slice $z$ (i.e., $x_z$) and the contextual information from both the $z^{+}$ and $z^{-}$ directions. One layer of CLSTM will integrate the information from the $z^{-}$-direction (resp., $z^{+}$-direction) and $x_z$ to capture the minus-side (resp., plus-side) context (see Fig.~\ref{fig:bdclstm}(B)). Then, the two one-side contexts ($z^{+}$ and $z^{-}$) will be fused. 

In fact, Pyramid-LSTM \cite{pyramid} can be viewed as a different extension of CLSTM, which employs six CLSTMs in six different directions ($x^{+/-}$, $y^{+/-}$, and $z^{+/-}$) and sums up the outputs of the six CLSTMs. However, useful information may be lost during the output summation. Intuitively, the sum of six outputs can only inform a simplified context instead of the exact situations in different directions. It should be noted that concatenating six outputs may greatly increase the memory consumption, and is thus impractical in Pyramid-LSTM. Hence, besides avoiding problematic convolutions on the $xz$ and $yz$ planes (as discussed in Section~\ref{sec:intro}), BDC-LSTM is in principle more effective in exploiting inter-slice contexts than Pyramid-LSTM. 

 

\textbf{Deep Architectures}: Multiple BDC-LSTMs can be stacked into a deep structure by taking the output feature map of one BDC-LSTM as the input to another BDC-LSTM. In this sense, each BDC-LSTM can be viewed as a super ``layer" in the deep structure. Besides simply taking one output as another input, we can also insert other operations, like max-pooling or deconvolution, in between BDC-LSTM layers. As a consequence, deep architectures for 2D CNN can be easily migrated or generalized to build deep architectures for BDC-LSTM. This is shown in Fig.~\ref{fig:bdclstm}(C)-(D). The underlying relationship between deep BDC-LSTM and 2D deep CNN is that deep CNN extracts a hierarchy of non-linear features from a 2D image and a deeper layer aims to interpret higher level information of the image, while deep BDC-LSTM extracts a hierarchy of hierarchical contextual features from the 3D context and a deeper BDC-LSTM layer seeks to interpret higher level 3D contexts. 

In \cite{convLSTM,Shi,pyramid}, multiple CLSTMs were simply stacked one by one, maybe with different kernel sizes, in which a CLSTM ``layer'' may be viewed as a degenerated BDC-LSTM ``layer''. When considering the problem in the context of CNN, as discussed above, one can see that no feature hierarchy was even formed in these simple architectures. Usually, convolutional layers are followed by subsampling, such as max-pooling, in order to form the hierarchy.

We propose a deep architecture combining max-pooling, dropout and deconvolution layers with the BDC-LSTM layers. The detailed structure is as follows (the numbers in parentheses indicate the size changes of the feature map in each 2D slice). Input ($64\!\times\!126\!\times\!126$), dropout layer with $p\!=\!0.5$, two BDC-LSTMs with 64 hidden units and $5\!\times\!5$ kernels ($64\!\times\!118\!\times\!118$), $2\!\times\!2$ max-pooling ($64\!\times\!59\!\times\!59$), dropout layer with $p\!=\!0.5$, two BDC-LSTMs with 64 hidden units and $5\!\times\!5$ kernels ($64\!\times\!51\!\times\!51$), $2\!\times\!2$ deconvolution ($64\!\times\!102\times\!102$), dropout layer with $p\!=\!0.5$, $3\!\times\!3$ convolution layer without recurrent connections ($64\!\times\!100\!\times\!100$), $1\!\times\!1$ convolution layer without recurrent connections ($2\!\times\!100\!\times\!100$). (Note: All convolutions in BDC-LSTM use the same kernel size as indicated in the layers.) Thus, to predict the probability map of a $100\!\times\!100$ region, we need the $126\!\times\!126$ region centered at the same position as the input. In the evaluation stage, the whole feature map can be processed using the overlapping-tile strategy \cite{unet}, because deep BDC-LSTM is fully convolutional along the $z$-direction. Suppose the feature map of a whole slice is of size $64\!\times\!W\!\times\!H$. The input tensor will be padded with zeros on the borders to resize into $64\!\times\!(W\!+\!26)\!\times\!(H\!+\!26)$. Then, a sequence of $64\!\times\!126\times\!126$ patches will be processed each time. The results are stitched to form the 3D segmentation.

\subsection{Combining $k$U-Net and BDC-LSTM}

The motivation of solving 3D segmentation by combining FCN ($k$U-Net) and RNN (BDC-LSTM) is to distribute the burden of exploiting 3D contexts. $k$U-Net extracts and compresses the hierarchy of intra-slice contexts into feature maps, and BDC-LSTM distills the 3D context from a sequence of abstracted 2D contexts. These two components work coordinately, as follows.

Suppose the 3D image consists of $N_z$ 2D slices of size $N_x\times N_y$ each. First, $k$U-Net extracts feature maps of size $64\times N_x \times N_y$, denoted by $f_{2D}^z$, from each slice $z$. The overlapping-tile strategy \cite{unet} will be adopted when the 2D images are too big to be processed by $k$U-Net in one shot. Second, BDC-LSTM works on $f_{2D}^z$ to build the hierarchy of non-linear features from 3D contexts and generate another $64\times N_x \times N_y$ feature map, denoted by $f_{3D}^{z}$, $z=1, \ldots N_z$. For each slice $z$, $f_{2D}^{h}$ ($h=z\!-\!\rho, \ldots, z, \ldots, z\!+\!\rho$) will serve as the context ($\rho=1$ in our implementation). Finally, a softmax function is applied to $f_{3D}^z$ to generate the 3D segmentation probability map.

\subsection{Training Strategy}

Our whole network, including $k$U-Net and BDC-LSTM, can be trained either end-to-end or in a decoupled manner. Sometimes, biomedical images are too big to be processed as a whole. Overlapping-tile is a common approach \cite{unet}, but can also reduce the range of the context utilized by the networks. The decoupled training, namely, training $k$U-Net and BDC-LSTM separately, is especially useful in situations where the effective context of each voxel is very large. Given the same amount of computing resources (e.g., GPU memory), when allocating all resources to train one component only, both $k$U-Net and BDC-LSTM can take much larger tiles as input. In practice, even though the end-to-end training has its advantage of simplicity and consistency, the decoupled training strategy is preferred for challenging problems.

$k$U-Net is initialized using the strategy in \cite{init} and trained using Adam \cite{adam}, with \textit{first moment coefficient} ($\beta_1$)=0.9, \textit{second moment coefficient} ($\beta_2$)=0.999, $\epsilon$=$1\mathrm{e}{-10}$, and a constant learning rate $5\mathrm{e}{-5}$. The training method for BDC-LSTM is Rms-prop \cite{rmsprop}, with \textit{smoothing constant} ($\alpha$)=0.9 and $\epsilon$=$1\mathrm{e}{-5}$. The initial learning rate is set as $1\mathrm{e}{-3}$ and halves every 2000 iterations, until $1\mathrm{e}{-5}$. In each iteration, one training example is randomly selected. The training data is augmented with rotation, flipping, and mirroring. To avoid gradient explosion, the gradient is clipped to $[-5,5]$ in each iteration. The parameters in BDC-LSTM are initialized with random values uniformly selected from $[-0.02, 0.02]$.

We use a weighted cross-entropy loss in both the $k$U-Net and BDC-LSTM training. In biomedical image segmentation, there may often be certain important regions in which errors should be reduced as much as possible. For instance, when two objects touch tightly to each other, it is important to make correct segmentation along the separating boundary between the two objects, while errors near the non-touching boundaries are of less importance. Hence, we adopt the idea in \cite{unet} to assign a unique weight for each voxel in the loss calculation. 

\section{Experiments}

Our framework was implemented in Torch7 \cite{torch} and the RNN package \cite{torchRNN}. We conducted experiments on a workstation with 12GB NVIDIA TESLA K40m GPU, using CuDNN library (v5) for GPU acceleration. Our approach was evaluated in two 3D segmentation applications and compared with several state-of-the-art DL methods.

\begin{table}[t]
  \caption{Experimental results on the ISBI neuron dataset and in-house 3D fungus datasets.}
  \label{tab:exp}
  \centering
  \begin{tabular}{lccc}
    \toprule
     & \multicolumn{2}{c}{Neuron} &  \multicolumn{1}{c}{Fungus} \\
    \cmidrule(l{2pt}r{2pt}){2-3}   \cmidrule(l{2pt}r{2pt}){4-4}
    Method    & $V_{rand}$     & $V_{info}$ &   Pixel Error  \\
    \midrule
    Pyramid-LSTM \cite{pyramid}       &  0.9677  &   0.9829   &  N/A\\
    U-Net \cite{unet}                           &  0.9728  &  0.9866      &  0.0263 \\
    Tri-Planar \cite{triplane}                &  0.8462   & 0.9180     &  0.0375 \\
    3D Conv \cite{3Dconv}		       &  0.8178   & 0.9125    & 0.0630 \\
    \midrule
    Ours (FCN only)                    & 0.9749  & 0.9869   &    0.0242 \\
    Ours (FCN+simple RNN)                       & 0.9742  & 0.9869   &   0.0241 \\
    Ours (FCN+deep RNN)                             & \textbf{0.9753}  & \textbf{0.9870}   &  \textbf{0.0215}   \\
    \bottomrule
  \end{tabular}
\end{table}

\textbf{3D Neuron Structures}: The first evaluation dataset was from the ISBI challenge on the segmentation of neuronal structures in 3D electron microscopic (EM) images \cite{ISBI_EM}. The objective is to segment the neuron boundaries. Briefly, there are two image stacks of $512\times512\times30$ voxels, where each voxel measures $4\times4\times50\mu m $. Noise and section alignment errors exist in both stacks. One stack (with ground truth) was used for training, and the other was for evaluation. We adopted the same metrics as in \cite{ISBI_EM}, i.e., foreground-restricted rand score ($V_{rand}$) and information theoretic score ($V_{info}$) after border thinning. As shown in \cite{ISBI_EM}, $V_{rand}$ and $V_{info}$ are good approximation to the difficulty for human to correct the segmentation errors, and are robust to border variations due to the thickness.
   
\textbf{3D Fungus Structures}: Our method was also evaluated on in-house datasets for the segmentation of tubular fungus structures in 3D images from Serial Block-Face Scanning Electron Microscope. The ratio of the voxel scales is $x:y:z=1:1:3.45$. There are five stacks, in all of which each slice is a grayscale image of $853\times 877$ pixels. We manually labeled the first 16 slices in one stack as the training data and used the other four stacks, each containing 81 sections, for evaluation. The metric to quantify the segmentation accuracy is {\it pixel error}, defined as the Euclidean distance between the ground truth label (0 or 1) and segmentation probability (a value in the range of $[0,1]$). Note that we do not use the same metric as the neuron dataset, because the ``border thinning" is not applicable to the fungus datasets. The pixel error was actually adopted at the time of the ISBI neuron segmentation challenge, which is also a well-recognized metric to quantify pixel-level accuracy. It is also worth mentioning that it is impractical to label four stacks for evaluation due to intensive labor. Hence, we prepared the ground truth every 5 sections in each evaluation stack (i.e., 5, 10, 15, $\ldots$, 75, 80). Totally, 16 sections were selected to estimate the performance on a whole stack. Namely, all 81 sections in each stack were segmented, but 16 of them were used to compute the evaluation score in the corresponding stack. The reported performance is the average of the scores for all four stacks.

Recall the four categories of known deep learning based 3D segmentation methods described in Section~\ref{sec:intro}. We selected one typical method from each category for comparison. (1) U-Net \cite{unet}, which achieved the state-of-the-art segmentation accuracy on 2D biomedical images, is selected as the representative scheme of linking 2D segmentations into 3D results. (Note: We are aware of the method \cite{AAAI} which is another variant of 2D FCN and achieved excellent performance on the neuron dataset. But, different from U-Net, the generality of \cite{AAAI} in different applications is not yet clear. Our test of \cite{AAAI} on the in-house datasets showed an at least $5\%$ lower F1-score than U-Net. Thus, we decided to take U-Net as the representative method in this category.) (2) 3D-Conv \cite{3Dconv} is a method using CNN with 3D convolutions. (3) Tri-planar \cite{triplane} is a classic solution to avoid high computing costs of 3D convolutions, which replaces 3D convolution with three 2D convolutions on orthogonal planes. (4) Pyramid-LSTM \cite{pyramid} is the best known generalized LSTM networks for 3D segmentation. 

\textbf{Results}: The results on the 3D neuron dataset and the fungus datasets are shown in Table~\ref{tab:exp}. It is evident that our proposed $k$U-Net, when used alone, achieves considerable improvement over U-Net \cite{unet}.  Our approach outperforms the known DL methods utilizing 3D contexts. Moreover, one can see that our proposed deep architecture achieves better performance than simply stacking multiple BDC-LSTMs together. As discussed in Section~\ref{sec:rnn}, adding subsampling layers like in 2D CNN makes the RNN component able to perceive higher level 3D contexts. It worth mentioning that our two evaluation datasets are quite representative. The fungus data has small anisotropism ($z$ resolution is close to $xy$ resolution). The 3D neuron dataset has large anisotropism ($z$ resolution is much less than $xy$ resolution). The effectiveness of our framework on handling and leveraging anisotropism can be demonstrated. 

We should mention that we re-implemented Pyramid-LSTM \cite{pyramid} in Torch7 and tested it on the fungus datasets. But, the memory requirement of Pyramid-LSTM, when implemented in Torch7, was too large for our GPU. For the original network structure, the largest possible cubical region to process each time within our GPU memory capacity was $40\times40\times8$. Using the same hyper-parameters in \cite{pyramid}, we cannot obtain acceptable results due to the limited processing cube. (The result of Pyramid-LSTM on the 3D neuron dataset was fetched from the ISBI challenge leader board\footnote{\url{http://brainiac2.mit.edu/isbi_challenge/leaders-board-new}} on May 10, 2016.) Here, one may see that our method is much more efficient in GPU memory, when implemented under the same deep learning framework and tested on the same machine. 

\begin{figure}[tb]
   \centering
   \includegraphics[width = 5.5in]{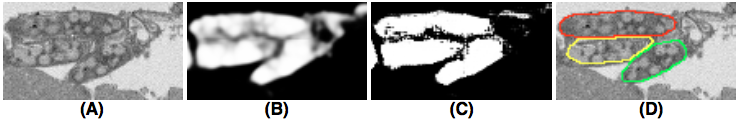} 
   \caption{(A) A cropped region in a 2D fungus image. (B) The result using only the FCN component. (C) The result of combining FCN and RNN. (D) The true fungi to be segmented in (A).}
   \label{fig:visual}
\end{figure}

Some results are shown in Fig.~\ref{fig:visual} to qualitatively compare the results using the FCN component alone and the results of combining RNN and FCN. In general, both methods make nearly no false negative errors. But, the RNN component can help to (1) suppress false positive errors by maintaining inter-slice consistency, and (2) make more confident prediction in ambiguous cases by leveraging the 3D context. In a nutshell, FCN collects as much discriminative information as possible within each slice and RNN makes further refinement according to inter-slice correlation, so that an accurate segmentation can be made at each voxel.

\section{Conclusions and Future Work}

In this paper, we introduce a new deep learning framework for 3D image segmentation, based on a combination of an FCN (i.e., $k$U-Net) to exploit 2D contexts and an RNN (i.e., BDC-LSTM) to integrate contextual information along the $z$-direction. Evaluated in two different 3D biomedical image segmentation applications, our proposed approach can achieve the state-of-the-art performance and outperform known DL schemes utilizing 3D contexts. Our framework provides a new paradigm to migrate the superior performance of 2D deep architectures to exploit 3D contexts. 
Following this new paradigm, we will explore BDC-LSTMs in different deep architectures to achieve further improvement and conduct more extensive evaluations on different datasets, such as BraTS (http://www.braintumorsegmentation.org/) and MRBrainS (http://mrbrains13.isi.uu.nl). 

\section{Acknowledgement}
This research was support in part by NSF Grants CCF-1217906 and CCF-1617735 and NIH Grants R01-GM095959 and U01-HL116330. Also, we would like to thank Dr.~Viorica Patraucean at University of Cambridge (UK) for discussion of BDC-LSTM, and Prof.~David P. Hughes and Dr.~Maridel Fredericksen at Pennsylvania State University (US) for providing the 3D fungus datasets.

\bibliographystyle{abbrv}
{\small
\bibliography{main}}

%
%
%
%
%

\end{document}